# Recognition of Documents in Braille

Dr Jomy John
*Assistant Professor*
Department of Computer Science
K. K. T. M. Govt. College, Pullut, Kodungallur, Kerala
jomyeldos@gmail.com

## 1 INTRODUCTION

Our society is diverse. There are people around us with vision impairment either partially or fully and people with hearing loss, people who unable to speak, people with inherent learning disability etc. Visually impaired people are integral part of the society and it has been a must to provide them with means and system through which they may communicate with the world. As of 2012 there were 285 million people who were visually impaired of which 246 million had low vision and 39 million were blind [1-2].

In this chapter, I would like to address how computers can be made useful to read the scripts in Braille. The importance of this work is to reduce communication gap between visually impaired people and the society. Braille remains the most popular tactile reading code even in this century. There are numerous amount of literature locked up in Braille. Their literary work, graphics are hidden under a script that normally others will not read. This work will help in office automation with huge documents held up in Braille. Braille recognition not only reduces time in reading or extracting information from Braille document but also helps people engaged in special education for correcting papers and other school related works. The availability of such a system will enhance communication and collaboration possibilities with visually impaired people [3].

Even if blind people know how to use Braille script, it is not an easy method to communicate with normal people because they are Braille illiterate. In addition to this, sufficient skilled trainers who could teach Braille are not available. The person who lost sight in a later stage found difficulty in learning this script.

To write the document, visually impaired people need a paper with high thickness. What they usually do is to combine two-three ordinary paper and write the document with a stylus. The embossed side is touched with their fingers to identify the letter written. This side needn't be a plain paper for them to read unlike ordinary people. Existing works supports only documents in white either bright or dull in colour. Hardly any work could be traced on hand printed ordinary documents in Braille.

This chapter is divided into four sections, Section II covers Braille writing system, Section III deals with materials and methods, Section IV covers results and discussion and Section V concludes the chapter.

## 2 BRAILLE WRITING SYSTEM

Braille is named after its creator, Frenchman Louis Braille, who lost his eyesight due to a childhood accident. Braille developed his code for the French alphabet in 1824. It was developed as an improvement on the night writing system developed by Charles Barbier. Night writing or sonography is a form of embossed writing, designed by Charles Barbier in response to Napoleon's demand for a code that soldiers could use to communicate silently and without light at night [4].

Braille is a writing system perceptible by touch and used by people who are blind or visually impaired. It is traditionally written with embossed paper. But in practice people used to write the script combining two or three normal paper together. Even the printed or handwritten pages are used by visually impaired people for writing because what normal people see is different from what blind people see. Braille users can also read computer screens that have refreshable Braille displays. They can write Braille with the original slate and stylus or type it on a Braille writer, such as a portable Braille note-taker or on a computer that prints with a Braille embosser. The embossed form of writing enables them to read the text without assistance. The slate and stylus is a quick, easy, convenient and constant method of making embossed printing for Braille character encoding. The writings are on the basis of six key points three on the left and three on the right as shown in Figure 1. The code for English and Malayalam is displayed in Figure 2 and 4 respectively. Figure 3 shows the grade 2 contractions.

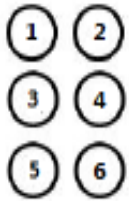

**Figure 1: Braille Cell**

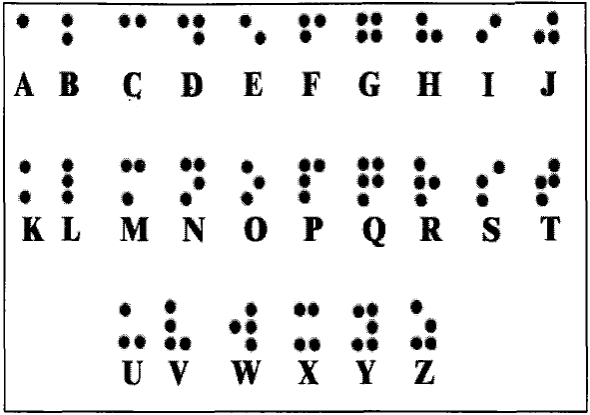

Figure 2: Braille codes for English Alphabets

| about | ⠁⠃⠕⠥⠞ | ab | ⠁⠃ |
|---|---|---|---|
| already | ⠁⠇⠗⠑⠁⠙⠽ | alr | ⠁⠇⠗ |

Figure 3: Grade 2 Contractions

Figure 4: Braille codes for Malayalam alphabets

Braille characters are small rectangular blocks called cells that contain tiny palpable bumps called raised dots. The number and position of these dots distinguish one character from another. Since the a variety of Braille alphabets originated as transcription codes of printed writing systems, the mappings vary from one language to another language. Furthermore, in English Braille it can be of three levels of encoding: Grade 1 is widely used as a letter-by-letter transcription used for basic literacy; Grade 2, is used an addition of abbreviations and contractions; and Grade 3, various non standardized personal shorthands [5]. For regional languages, like Malayalam Bharathi Braille is used which adopts Unicode method.

Braille cells are not the only thing to appear in Braille text. There may be embossed illustrations and graphs, with the lines either solid or made of series of dots, arrows, bullets that are larger than Braille dots, etc. A full Braille cell includes six raised dots arranged in two lateral rows each having three dots. The dot positions are identified by numbers from one through six. 64 solutions are possible from using one or more dots. A single cell can be used to represent an alphabet letter, number, punctuation mark, or even an entire word.

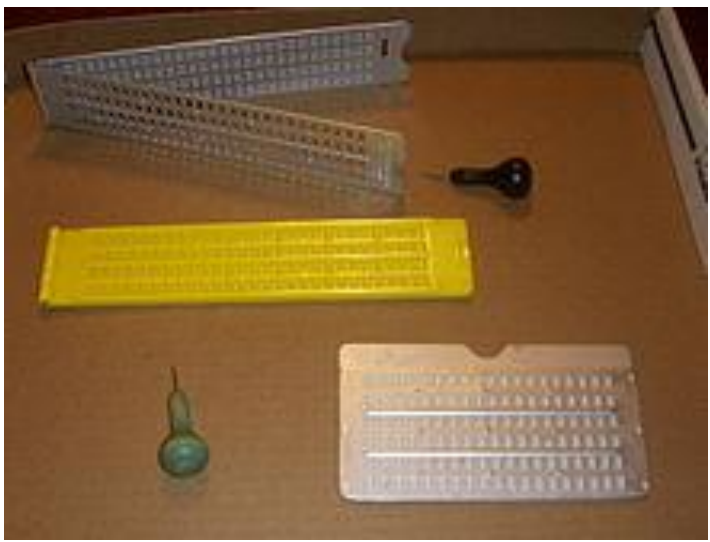

**Figure 5: Different types of stylli and slate**

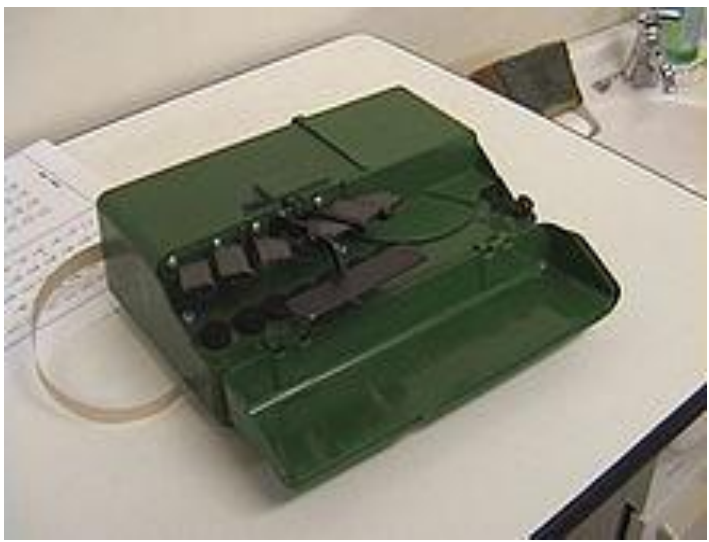

**Figure 6: Braille typewriter**

Because Braille letters cannot be effectively erased and written over if an error is made, an error is overwritten with all six dots (⠿). The basic design of the slate consists of two pieces of metal, plastic or wood fastened together with a hinge at one side. The stylus is a short blunted awl with a handle to comfortably fit the hand of the user.

Writing is accomplished by placing a piece of heavy paper in the slate, aligning it correctly and closing the slate. The pins in the back of the slate puncture or pinch the paper securely between the two halves of the slate. The person's writing begins in the upper right corner with each combination of dots in the cell written backward. The awl is positioned and pressed to form a depression in the paper. The writer moves to one of the other dots in the cell or to the next cell as appropriate. The slate is repositioned as needed to continue writing on the paper. When completed the writer removes the slate and turns the paper over to read the Braille by feeling the dots that were pushed up from the back. Braille embossers usually need special Braille paper which is thicker and more expensive than normal paper. Some high-end embossers are capable of printing on normal paper. Embossers can be either one-sided or two-sided. Duplex embossing requires lining up the dots so they do not overlap known as inter-point. Multiple copies of a Braille document can be produced using a device called thermoform but lacks the original quality.

## 3 MATERIALS AND METHODS

In this chapter, the recognition of English and Malayalam documents in Braille are considered. Grade 1 Braille is taken care of in the case of English documents. Only the vowels are taken into consideration for the present work. The recognizer includes steps such as image acquisition, pre-processing, feature extraction and template matching. Figure 7 shows the frame work of the proposed system. The slate and stylus used in the present work are shown in figure 8. Figure 9 shows the Braille codes in printed as well as plain paper.

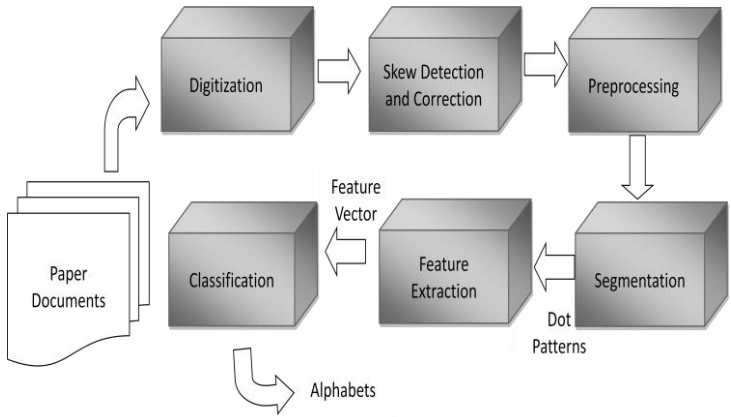

Figure 7: Framework of the proposed system

**225**

**Image Acquisition**: Using a flat-bed scanner, images of single sided embossed Braille documents is taken. Without the need to carry out complex modifications, the scanner can be used with any other application.

**Image Pre-processing**: In Pre-processing, first of all conversion of RGB image to Gray-scale image is done. Formula for covert RGB to gray level is as follow $gray = 0.2989\ R + 0.5870\ G + 0.1140\ B$ [6]

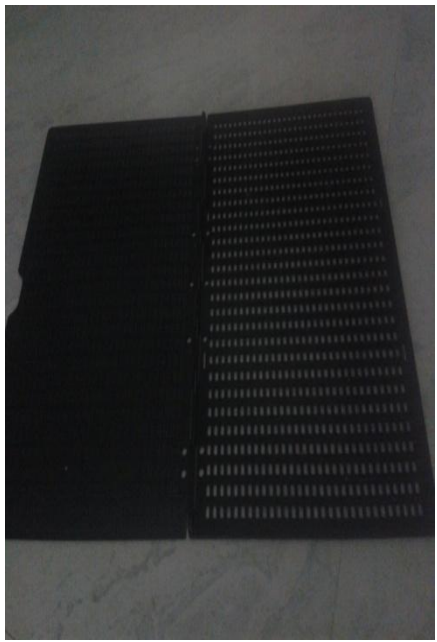

Figure 8: (a) Slate

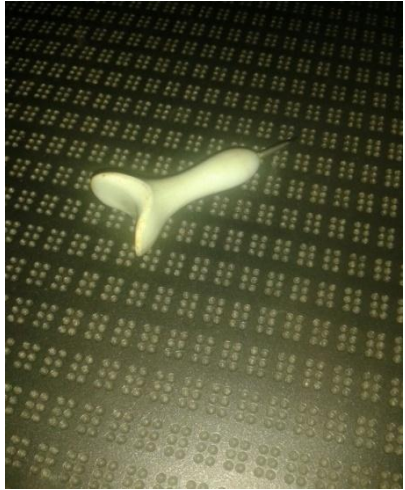

**Figure 8: (a) Stylus**

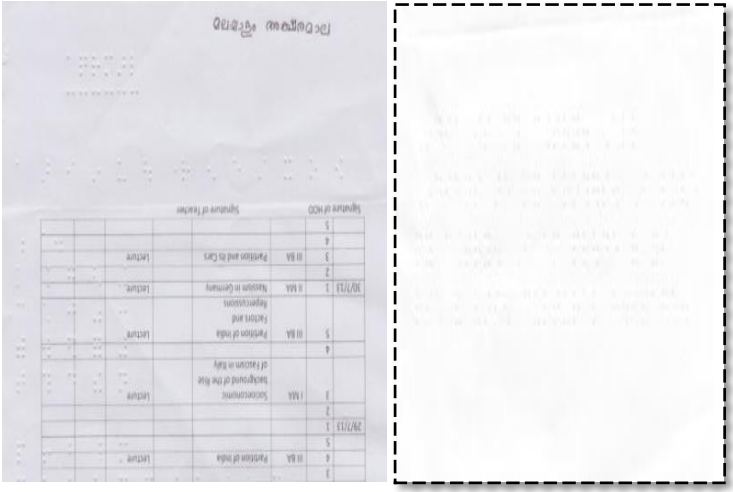

**Figure 9: Braille codes on printed papers and plain paper**

**Image enhancement:** To highlight specific image features is the main goal of image enhancement. The main features of image are the dots and their relative location. Few image enhancement techniques are performed to make these dots bold and easier to localize in subsequent steps.

**Noise Reduction**: Noise is eliminated by average filter.

**Contrast Enhancement:** To concentrate on the intensity range around the dots intensity levels, contrast stretching method is used.

**Image Segmentation:** To separate the desired dots from the background image segmentation is used. Image compliment is performed to obtain well-defined dots, after that image dilation is performed to dilate the dots. At last binarization step is performed to separate the dots from the background.

**Feature Extraction**: For each cell, the presence of a dot is detected and coded as 1 and absence of a dot is coded as 0. In this phase extraction of the relevant information from the image for interpretation of the letters and words is performed.

**Braille Cells Recognition**: According to the features extracted in the previous stage Braille cell

recognition can be done. Major step of this phase is to compose meaningful letters. It aims to obtain letters and words by grouping the dots based on the location information. So to form words cell recognition is necessary

## 4 RESULTS AND DISCUSSION

The recognizer for Braille documents in grade 1 English and Malayalam vowels are presented. Figure 10 shows the segmented line containing Malayalam vowels. Figure 11 shows the enhanced image of figure 10. Figure 12 displays the image after binarization. Table 1 shows the coding strategy for Malayalam vowels. The recognition result is displayed in Figure 13 for Malayalam vowels and Figure 14 displays the result for English Braille.

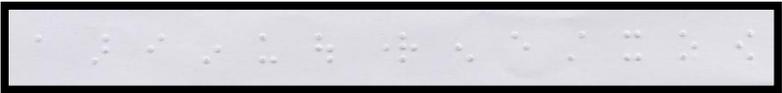

**Figure 10: Malayalam Vowels**

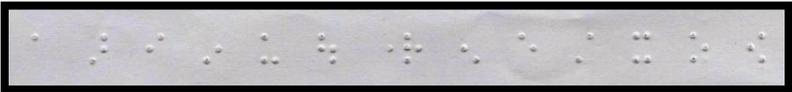

**Figure 11: Enhanced Malayalam Vowels**

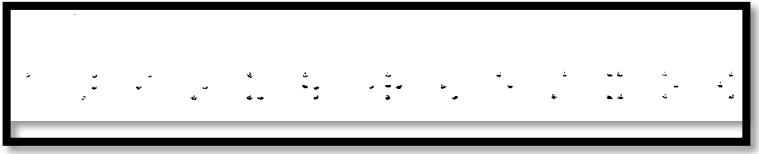

Figure 12: Effect of Binarization

| | | | | | | | |
|---|---|---|---|---|---|---|---|
| അ | 100000 | ഉ | 100011 | ഒ | 010010 |
| ആ | 010110 | ഊ | 101101 | ഒെ | 010010 |
| ഇ | 011000 | എ | 001001 | ഓ | 100110 |
| ഈ | 000110 | ഏ | 100100 | ഔ | 011001 |

Table 1: Coding strategy for Malayalam vowels

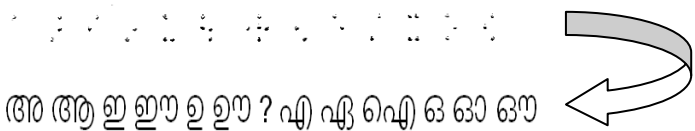

Figure 13: The recognition result for Malayalam vowels

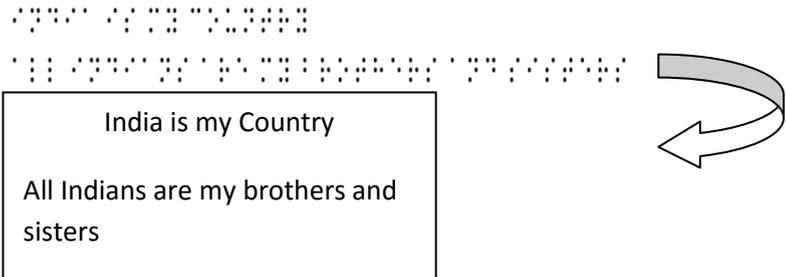

India is my Country

All Indians are my brothers and sisters

Figure 14: The recognition result for English document

## 4 CONCLUSION

In this chapter, I have addressed the problem of the recognition of hand printed documents in Braille for English as well as Malayalam. In the case of Malayalam only vowels were considered. The code including conjunct characters, symbols, etc remain as future work. This work can be extended using Grade 2 or Grade 3 Braille. The documents are captured through scanner in this work. But the digital camera or mobile camera captured work also could be incorporated.  Braille script in bilingual language is another issue to be taken care of. The ultimate aim in recognizing Malayalam Braille is to recognize scripts like the one displayed below.

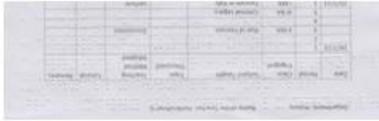

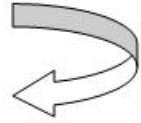

അമ്മയും നന്മയും ഒന്നാണ്
ഞങ്ങളും നിങ്ങളും ഒന്നാണ്
അറ്റമില്ലാത്തൊരുജീവിതപാതയില്‍
ഒറ്റയല്ലൊറ്റയല്ലൊറ്റയല്ല

## ACKNOWLEDGEMENT

I would like to thank Mr. G. Harikrishnan, Assistant Professor in History, K. K. T. M. Government College for spending time for writing the documents in English and Malayalam and for providing me all the help and support for making me understand how they read and write Braille.